\documentclass[conference]{IEEEtran}
\IEEEoverridecommandlockouts
\usepackage{cite}
\usepackage{amsmath,amssymb,amsfonts}
\usepackage{algorithmic}
\usepackage{graphicx}
\usepackage{textcomp}
\usepackage{xcolor}
\def\BibTeX{{\rm B\kern-.05em{\sc i\kern-.025em b}\kern-.08em
    T\kern-.1667em\lower.7ex\hbox{E}\kern-.125emX}}
\begin{document}

\title{Intracranial Hemorrhage Detection Using Neural Network Based Methods With Federated Learning*\\

}

\author{\IEEEauthorblockN{1\textsuperscript{st} Utkarsh Chandra Srivastava}
\IEEEauthorblockA{\textit{dept. Computer Science Engineering} \\
\textit{SRMIST Kattankulathur, Chennai}\\
uu7127@srmist.edu.in,}
\and
\IEEEauthorblockN{2\textsuperscript{nd} Anshuman Singh }
\IEEEauthorblockA{\textit{dept. Computer Science Engineering} \\
\textit{SRMIST Kattankulathur, Chennai}\\
as4917@srmist.edu.in}
\and
\IEEEauthorblockN{3\textsuperscript{nd} Dr. K. Sree Kumar}
\IEEEauthorblockA{\textit{dept. Computer Science Engineering} \\
\textit{SRMIST Kattankulathur, Chennai}\\
 sreekumar.k@ktr.srmuniv.ac.in}

}

\maketitle

\begin{abstract}
Intracranial hemorrhage, bleeding that occurs inside the cranium, is a serious health problem requiring rapid and often intensive medical treatment. Such a condition is traditionally diagnosed by highly-trained specialists analyzing computed tomography (CT) scan of the patient and identifying the location and type of hemorrhage if one exists. We propose a neural network approach to find and classify the condition based upon the CT scan.  The model architecture implements a time distributed convolutional network\cite{b3}. We observed accuracy above 92\% from such an architecture, provided enough data. We propose further extensions to our approach involving the deployment of federated learning\cite{b4}. This would be helpful in pooling learned parameters without violating the inherent privacy of the data involved.
\end{abstract}


\section{Introduction}
When a patient shows acute neurological symptoms such as severe headache or loss of consciousness, highly trained specialists review medical images of the patient’s cranium to look for the presence, location and type of hemorrhage. The process is complicated and often time-consuming as these scans are essentially a combination of multiple X-Ray scans processed by a computer. To reduce delays that lead to deaths we propose a solution to this problem based on a deep learning approach to automate the detection of intracranial hemorrhaging. We employ a modified version of DenseNet1212 trained on a dataset provided by the Radiological Society of North America. These convolutional networks are then stacked using a sequence model (such as a recurrent neural network) With GRUs\cite{b2} to preserve temporal information. The system can detect acute intracranial hemorrhage and its subtypes with accuracy greater than 92\%. We aim to provide automated, faster and more accurate diagnosis using computer vision and deep learning frameworks for image recognition and classification. Our work here aims to reduce the computational load of the model via pruning and repurposing the architecture to create a more practical model for inference at even remote locations with limited computation power.  

\section{Methods}

\subsection{Architecture Of Our Model}

\begin{figure}[htbp]
\centerline{\includegraphics[scale=0.3]{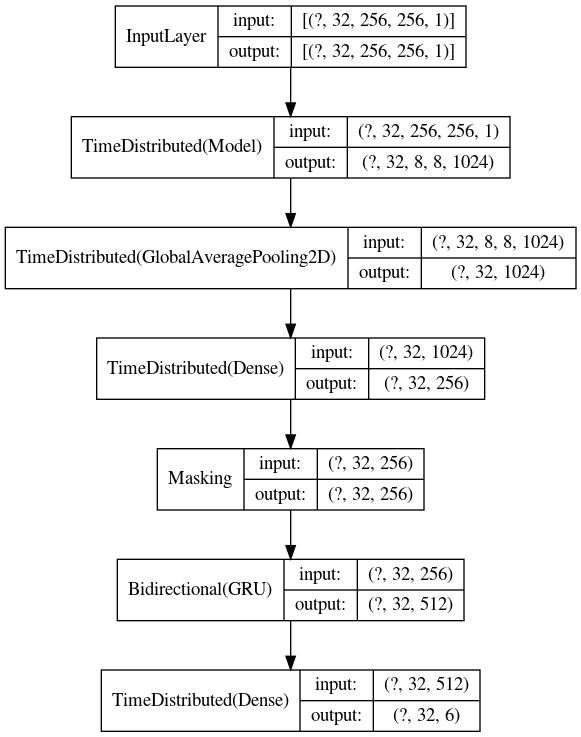}}
\caption{Model Architecture}
\label{arch}
\end{figure}
   1. Convolution Pass [DenseNet 121] :\newline
DenseNet\cite{b1} architecture is new, it is a logical extension of ResNet. They alleviate the vanishing gradient problem, strengthen feature propagation, encourage feature reuse, and substantially reduce the number of parameters.

    2. Our recurrent network :\newline
Our recurrent network builds upon the DenseNet121 model but adding a ‘TimeDistributedModel’ layer to the output of the DenseNet Model. This layer serves to collect temporal correlations from the multiple X-Ray images of a single CT Scan. This allows us to better predict the labels from our data as the temporal evolution of the brain is accounted for.

    3. Federated Pass :
In order to train on the data which may include highly sensitive medical data. The federated pass aggregates the weights of models trained on the device and then updates the model output. Hence we can train a common model without collecting data.
\section{Results And Discussion}

\subsection{Model Performance}

\begin{figure}[htbp]
\centerline{\includegraphics[scale=0.25]{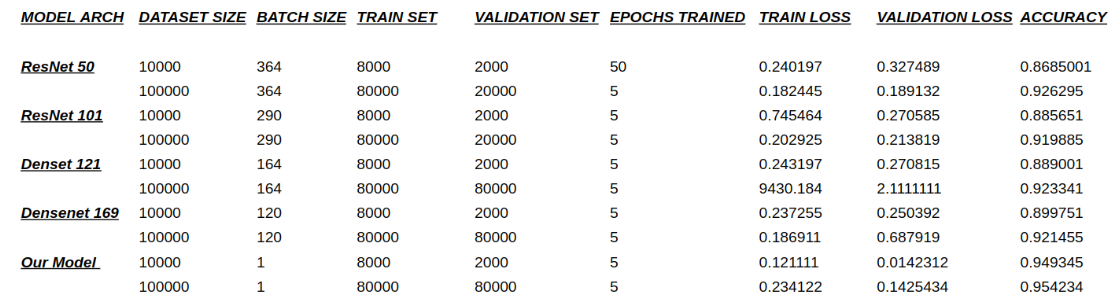}}
\caption{Model Benchmark}
\label{arch}
\end{figure}
 
We benchmarked our approach against the following existing traditional Convolutional Models:

    1) ResNet 50 
    2) ResNet 101 
    
    3) DenseNet 121
    4) DenseNet 169

There are further advancements that can be made in our approach to boost the accuracy and the mean average precision score of our model but being limited to the computational constraints while during our research were not able to validate our expectations.Further research advancements in our approach involve the implementation of Federated Learning5 and using 3D-Convolutions6 for our convolution step instead of traditional 2D-Convolutions.

\begin{figure}[htbp]
\centerline{\includegraphics[scale=0.34]{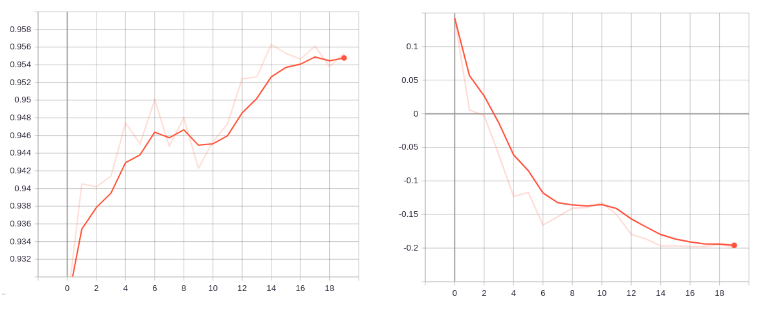}}
\caption{Accuracy And Loss}
\label{archfed}
\end{figure}

\subsection{Federated Learning}

Federated learning, introduced by Google in 2017, is a distributed machine learning approach that enables multi-institutional collaboration on deep learning projects without sharing patient data. 

Federated learning will bring AI with privacy to hospitals with techniques like model encryption only the model updates are shared with the central model aggregator. This provides protection to both the model and the data. The raw data never leaves the institutions, which not only adds privacy but also prevents large data transfers on the network.

The rising popularity of edge computing will give a boost to research, development, and adoption of Federated Learning. In the years to come, Federated Learning will become an indispensable tool to carry out privacy-preserving, distributed learning on decentralized data.

\begin{figure}[htbp]
\centerline{\includegraphics[scale=0.6]{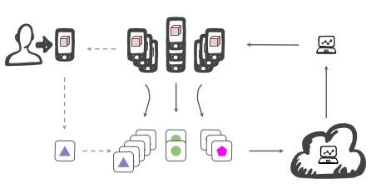}}
\caption{Federated Step}
\label{archfed}
\end{figure}

\subsection{Challenges}
\begin{itemize}

\item The data from the hospitals can be extremely skewed, depending on the specialty of the doctors in question. This works in our favor as it allows us to take advantage of the federated passes and better personal inference scenarios.

\item We also note that a federated environment with Non-IID data will pose more challenges to the training paradigm of the model. 

\item Expensive Communication: Federated networks are potentially comprised of a massive number of devices (e.g., thousands of hospitals), and communication in the network can be slower than local computation by many orders of magnitude.

\item Limited storage: and compute resources on the client: The clients are typically light in storage and compute power. Especially when the clients are mobile devices, it is important that the FL implementation kicks in only when the device is available (it is plugged in and on an unmetered Internet connection

\end{itemize}

\begin{figure}[htbp]
\centerline{\includegraphics[scale=0.21]{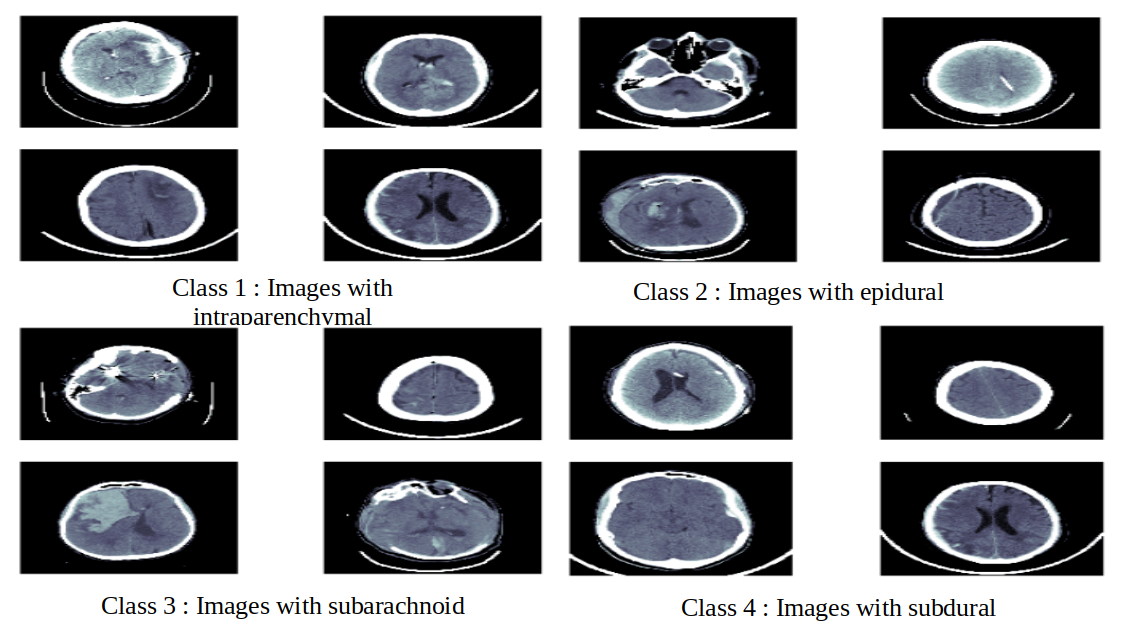}}
\caption{Types Of Hemorrhage Classified}
\label{arch}
\end{figure}

\subsection{Further Research Possibilities}

\subsubsection{Differential Privacy\cite{b4}}

The encrypted model is sent to the individual institutions  which decrypt within a secure enclave in hardware and then train on the local data. Only the model updates are shared with the central model aggregator. This provides protection to both the model and the data.

The raw data never leaves the institutions, which not only adds privacy and reinforce patient confidentiality but also prevents large data transfers on the network.

\subsubsection{3D-Convolutions\cite{b5}}

The 3D activation map produced during the convolution of a 3D CNN is necessary for analyzing data where temporal or volumetric context is important. This ability to analyze a series of frames or images in context has led to the use of 3D CNNs as tools for action recognition and evaluation of medical imaging.

\subsubsection{Faster Recurrent Networks\cite{b6}}
Deep recurrent neural networks (RNN), such as LSTM, have many advantages over forward networks. However, the LSTM training method, such as backward propagation through time (BPTT), is really slow. 
\section{Conclusion}

We find many real-world advantages to a deep learning-based system, over traditional workflows only involving humans. The system remains robust to a large variety of data and classes. The global model keeps updating due to the federated passes and retains accuracy throughout its lifespan.

\end{document}